\def\year{2018}\relax
\documentclass[letterpaper]{article} 
\usepackage{aaai18}  
\usepackage{times}  
\usepackage{helvet}  
\usepackage{courier}  
\usepackage{url}  
\usepackage{graphicx}  
\usepackage{latexsym}
\usepackage{amsmath}
\usepackage{amssymb}
\usepackage{xspace}
\usepackage{CJKutf8}
\usepackage{algorithm, algorithmicx, algpseudocode}
\usepackage{subcaption}
\usepackage{multirow,array}
\usepackage{makecell}
\usepackage{color}

\makeatother

\newcommand{\citet}[1]{\citeauthor{#1}~\shortcite{#1}}

\begin{document}

\title{Joint Training for Neural Machine Translation Models \\ with Monolingual Data}
\date{}

\author{ Zhirui Zhang$^\dag$, Shujie Liu$^\ddag$, Mu Li$^\ddag$, Ming Zhou$^\ddag$, Enhong Chen$^\dag$\thanks{Corresponding author} \\
  $^\dag$University of Science and Technology of China, Hefei, China\\
   $^\ddag$Microsoft Research \\
    $^\dag$zrustc11@gmail.com \  $^\dag$cheneh@ustc.edu.cn \\ 
    $^\ddag$\{shujliu,muli,mingzhou\}@microsoft.com \\
}

\begin{CJK*}{UTF8}{gbsn}

\maketitle

\begin{abstract}
Monolingual data have been demonstrated to be helpful in improving translation quality of both statistical machine translation (SMT) systems and neural machine translation (NMT) systems, especially in resource-poor or domain adaptation tasks where parallel data are not rich enough.
In this paper, we propose a novel approach to better leveraging monolingual data for neural machine translation by jointly learning source-to-target and target-to-source NMT models for a language pair with a joint EM optimization method.
The training process starts with two initial NMT models pre-trained on parallel data for each direction, and these two models are iteratively updated by incrementally decreasing translation losses on training data.
In each iteration step, both NMT models are first used to translate monolingual data from one language to the other, forming pseudo-training data of the other NMT model.
Then two new NMT models are learnt from parallel data together with the pseudo training data. 
Both NMT models are expected to be improved and better pseudo-training data can be generated in next step.
Experiment results on Chinese-English and English-German translation tasks show that our approach can simultaneously improve translation quality of source-to-target and target-to-source models, significantly outperforming strong baseline systems which are enhanced with monolingual data for model training including back-translation.

\end{abstract}

\section{Introduction}

Neural machine translation (NMT) performs end-to-end translation based on an encoder-decoder framework ~\cite{Kalchbrenner,cho2014learning,sutskever2014sequence,Bahdanau2014NeuralMT} and has obtained state-of-the-art performances on many language  pairs~\cite{luong2015effective,Sennrich2016NeuralMT,tu-EtAl:2016:P16-1,Wu2016GooglesNM}.
In the encoder-decoder framework, an encoder first transforms the source sequence into vector representations, based on which, a decoder generates the target sequence.
Such framework brings appealing properties over the traditional phrase-based statistical machine translation (SMT) systems ~\cite{Koehn2003StatisticalPT,chiang2007hierarchical}, such as little requirements for human feature engineering, or prior domain knowledge.
On the other hand, to train the large amount of parameters in the encoder and decoder networks, most NMT systems heavily rely on high-quality parallel data and perform poorly in resource-poor or domain-specific tasks.
Unlike bilingual data, monolingual data are usually much easier to collect and more diverse,
and have been attractive resources for improving machine translation models since 1990's when data-driven machine translation systems were first built.

Monolingual data play a key role in training SMT systems. Additional target monolingual data are usually required to train a powerful language model, which is an important feature of an SMT system's log-linear model. Using source-side monolingual data in SMT were also explored. \citet{Ueffing2007transductive} introduced a transductive semi-supervised learning method, in which source monolingual sentences are translated and filtered to build pseudo bilingual data, which are added to the original bilingual data to re-train the SMT model.

For NMT systems, \citet{gulcehre2015using} first tried both shallow and deep fusion methods to integrate an external RNN language model into the encoder-decoder framework.
The shallow fusion method simply linearly combines the translation probability and the language model probability, while the deep fusion method connects the RNN language model with the decoder to form a new tightly coupled network.  
Instead of introducing an explicit language model, ~\citet{Cheng2016SemiSupervisedLF} proposed an auto-encoder-based method which encodes and reconstructs monolingual sentences, in which source-to-target and target-to-source NMT models serve as the encoder and decoder respectively.

~\citet{Sennrich2016ImprovingNM} proposed back-translation for data augmentation as another way to leverage the target monolingual data. 
In this method, both the NMT model and training algorithm are kept unchanged, instead they employed a new approach to constructing training data.
That is, target monolingual sentences are translated with a pre-constructed machine translation system into source language, which are used as additional parallel data to re-train the source-to-target NMT model.
Although back-translation has been proven to be robust and effective, one major problem for further improvement is the quality of automatically generated training data from monolingual sentences. Due to the imperfection of machine translation system, some of the incorrect translations are very likely to hurt the performance of source-to-target model.

In this paper, we present a novel method for making extended usage of monolingual data from both source side and target side by jointly optimizing a source-to-target NMT model $A$ and a target-to-source NMT model $B$ through an iterative process.
In each iteration, these two models serve as helper machine translation systems for each other as in back-translation: 
$B$ is used to generated pseudo-training data for model $A$ with target-side monolingual data, and $A$ is used to generated pseudo-training data for model $B$ with source-side monolingual data. 
The key advantage of our new approach comparing with existing work is that the training process can be repeated to obtain further improvements because after each iteration both model $A$ and $B$ are expected to be improved with additional pseudo-training data. Therefore, in the next iteration, better pseudo-training data can be generated with these two improved models, resulting even better model $A$ and model $B$, so on and so forth.

To jointly optimize the two models in both directions, we design a new semi-supervised training objective, with which the generated training sentence pairs are weighted so that the negative impact of noisy translations can be minimized. Original bilingual sentence pairs are all weighted as 1, while the synthetic sentence pairs are weighted as the normalized model output probability. 
Similar to the post-processing step as described in \citet{Ueffing2007transductive}, our weight mechanism also plays an important role in improving the final translation performance.
As we will show in the paper, the overall iterative training process essentially adds a joint EM estimation over the monolingual data to the MLE estimation over bilingual data: the E-step tries to estimate the expectations of translations of the monolingual data, while the M-step updates model parameters with the smoothed translation probability estimation.  

Our experiments are conducted on NIST OpenMT's Chinese-English translation task and WMT's English-German translation task.
Experimental results demonstrate that our joint training method can significantly improve translation quality of both source-to-target and target-to-source models, compared with back-translation and other strong baselines.

\section{Neural Machine Translation}
In this section, we will first briefly introduce the NMT model used in our work. 
The NMT model follows the attention-based architecture proposed by~\citet{Bahdanau2014NeuralMT}, and it is implemented as an encoder-decoder framework with recurrent neural networks (RNN). RNN are usually implemented as Gated Recurrent Unit (GRU)~\cite{cho2014learning} (adopted in our work) or Long Short-Term Memory (LSTM) networks~\cite{hochreiter1997long}.
The whole architecture can be divided into three components: encoder, decoder and attention mechanism.

\subsubsection{Encoder}
The encoder reads the source sentence \( X = (x_1, x_2, ...\, , x_T ) \) and transforms it into a sequence of hidden states \( h = (h_1, h_2,...\, ,h_T)\), using a bi-directional RNN. 
At each time stamp \(t\), the hidden state \(h_t\) is defined as the concatenation of the forward and backward RNN hidden states \([\overrightarrow{h_t};\overleftarrow{h_t} ]\), where \(\overrightarrow{h_t} = \text{RNN}(x_t, \overrightarrow{h_{t-1}}) \), \( \overleftarrow{h_t} = \text{RNN}(x_t, \overleftarrow{h_{t+1}}) \). 

\subsubsection{Decoder}
The decoder uses another RNN to generate the translation \(Y = (y_1, y_2, ...\, , y_{T'})\) based on the hidden states \(h\) generated by the encoder. 
At each time stamp \(i\), the conditional probability of each word \(y_i\) from a target vocabulary \(V_y\) is computed by
\begin{equation}
\label{equ-softmax}
p(y_i|y_{<i},h) = g(y_{i-1},z_i,c_i),
\end{equation}
where \(z_i\) is the $i_{th}$ hidden state of the decoder, which is calculated conditioned on the previous hidden state \(z_{i-1}\), previous word \(y_{i-1}\) and the source context vector \(c_i\): 
\begin{equation}
z_i = \text{RNN}(z_{i-1},y_{i-1},c_i),
\end{equation}
where the source context vector \(c_i\) is computed by the attention mechanism. 

\subsubsection{Attention Mechanism} The context vector \(c_i\) is a weighted sum of the hidden states \((h_1, h_2,...\, ,h_T)\) with the coefficients \(\alpha_1, \alpha_2, ...\, , \alpha_T\) computed by 
\begin{equation}
\alpha_t = \frac{\exp{ (a(h_t,z_{i-1}) )}}{\sum_k{\exp{(a(h_k, z_{i-1}))}}}
\end{equation}
where \(a\) is a feed-forward neural network with a single hidden layer. 

\subsubsection{MLE Training}
NMT systems are usually trained to maximize the conditional log-probability of the correct translation given a source sentence with respect to the parameters \( \theta \) of the model:
\begin{equation}
\theta^* = \arg\max \limits_{\theta} \sum_{n=1}^{N}\sum_{i=1}^{|y^n|} \log{p(y_i^n| y_{<i}^n, x^n)}
\label{equ:MLE-loss}
\end{equation}
where \( N\) is size of the training corpus, and \(|y^n|\) is the length of the target sentence \(y^n\).

As with the most of deep learning models, the model parameters $\theta^*$ have to be learnt with fully labeled data, which means parallel sentence pairs $(x_i, y_i)$ in the machine translation task, while monolingual data cannot be directly applied to model training. 

\section{Joint Training for Paired NMT Models}

\begin{figure}[t] 
	\begin{center}
		\includegraphics[scale=0.56]{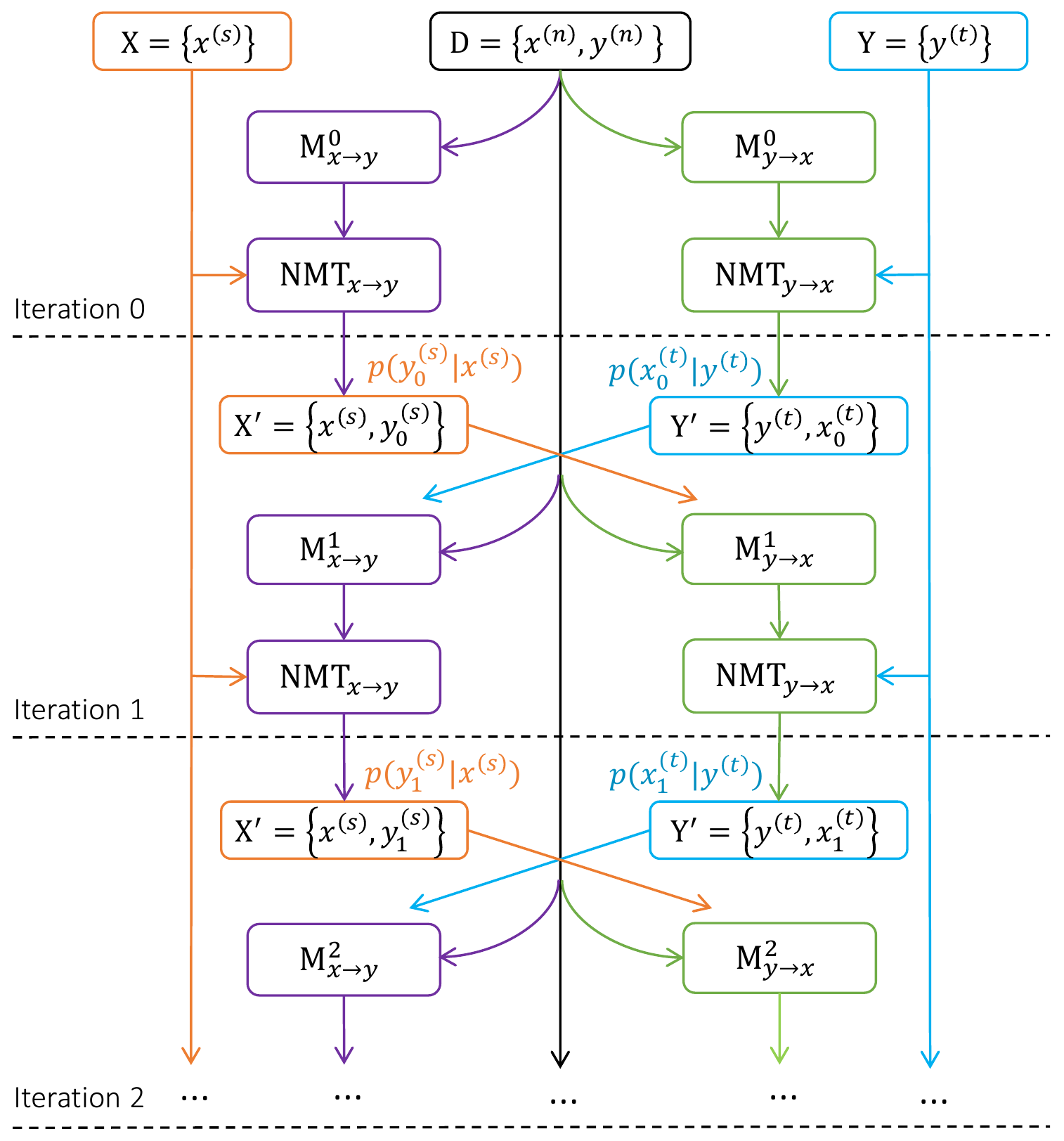}
		\caption{Illustration of joint-EM training of NMT models in two directions ($\text{NMT}_{x\rightarrow y}$ and $\text{NMT}_{y\rightarrow x}$) using both source ($X$) and target ($Y$) monolingual corpora, combined with bilingual data $D$. $X'$ is the generated synthetic data with probability $p(y|x)$ by translating $X$ using $\text{NMT}_{x\rightarrow y}$, and ${Y}'$ is the synthetic data with probability $p(x|y)$ by translating ${Y}$ using $\text{NMT}_{y\rightarrow x}$. }
		\label{fig:FrameworkOfJointTrain}
	\end{center}
\end{figure}
 
Back translation fills the gap between the requirement for parallel data and  availability of monolingual data in NMT model training with the help of machine translation systems.  
Specially, given a set of sentences $\{y_i\}$ in target language $Y$, a pre-constructed target-to-source machine translation system is used to automatically generate their translations $\{x_i\}$ in source language $X$. Then the synthetic sentence pairs $\{(x_i, y_i)\}$ are used as additional parallel data to train the source-to-target NMT model, together with the original bilingual data.
 
Our work follows this parallel data synthesis approach, but extends the task setting from solely improving the source-to-target NMT model training with target monolingual data to a paired one:
we aim to jointly optimize a source-to-target NMT model $M_{x\rightarrow y}$ and a target-to-source NMT model $M_{y\rightarrow x}$ with the aid of monolingual data from both source language $X$ and target language $Y$.
Different from back translation, in which both automatic translation and NMT model training are performed only once, our method runs the machine translation for monolingual data and updates NMT models $M_{x\rightarrow y}$ and $M_{y\rightarrow x}$ through several iterations. 
At each iteration step, model $M_{x\rightarrow y}$ and $M_{y\rightarrow x}$ serves as each other's pseudo-training data generator: $M_{y\rightarrow x}$ is used to translate $Y$ into $X$ for $M_{x\rightarrow y}$, while $M_{x\rightarrow y}$ is used to translate $X$ to $Y$ for $M_{y\rightarrow x}$. 
 
\begin{algorithm}[t]
\caption{Joint Training Algorithm for NMT}
\label{alg:jointTraining}
\begin{algorithmic}[1]
\Procedure{Pre-training}{}
\State Initialize $\text{M}_{x\rightarrow y}$ and $\text{M}_{y\rightarrow x}$ with random weights $\theta_{x\rightarrow y}$ and $\theta_{y\rightarrow x}$; 
\State Pre-train $\text{M}_{x\rightarrow y}$ and $\text{M}_{y\rightarrow x}$ on bilingual data $\text{D}=\{(x^{(n)},y^{(n)}\}_{n=1}^{N}$ with Equation~\ref{equ:MLE-loss};
\EndProcedure
\Procedure{Joint-training}{}
\While{Not Converged}
\State  Use $\text{NMT}_{y\rightarrow x}$ to generate back-translation $x$ for $\text{Y} = \{y^{(t)}\}_{t=1}^{T}$ and build pseudo-parallel corpora $\text{Y}' = \{x,y^{(t)}\}_{t=1}^{T}$; \Comment{E-Step for $\text{NMT}_{x\rightarrow y}$}
\State Use $\text{NMT}_{x\rightarrow y}$ to generate back-translation $y$ for $\text{X} = \{x^{(s)}\}_{s=1}^{S}$ and build pseudo-parallel corpora $\text{X}' = \{x^{(s)},y\}_{s=1}^{S}$;  \Comment{ E-Step for $\text{NMT}_{y\rightarrow x}$}
\State Train $\text{M}_{x\rightarrow y}$ with Equation~\ref{equ:loss_x2y} given weighted bilingual corpora $\text{D}\cup \text{Y}'$;   \Comment{ M-Step for $\text{NMT}_{x\rightarrow y}$}
\State Train $\text{M}_{y\rightarrow x}$ with Equation~\ref{equ:loss_y2x} given weighted bilingual corpora $\text{D}\cup \text{X}'$;   \Comment{M-Step for $\text{NMT}_{y\rightarrow x}$}
\EndWhile
\EndProcedure
\end{algorithmic}
\end{algorithm}

The joint training process is illustrated in Figure~\ref{fig:FrameworkOfJointTrain},
in which the first 2 iterations are shown. 
Before the first iteration starts, two initial translation models $M_{x\rightarrow y}^0$ and $M_{y\rightarrow x}^0$ are pre-trained with parallel data $D = \{x^n, y^n\}$.
This step is denoted as iteration 0 for sake of consistency.

In iteration 1, at first, two NMT systems based on $M_{x\rightarrow y}^0$ and $M_{y\rightarrow x}^0$ are used to translate monolingual data $X=\{x_i^{(s)}\}$ and $Y=\{y_i^{(s)}\}$, which forms two synthetic training data sets $X'=\{x_i^{(s)}, y_0^{(s)}\}$ and $Y'=\{y_i^{(t)}, x_0^{(t)}\}$. 
Model $M_{x\rightarrow y}^1$ and $M_{y\rightarrow x}^1$ are then trained on the updated training data by combining $Y'$ and $X'$ with parallel data $D$.
It is worth noting that we use n-best translations from an NMT system, and the selected translations are weighted with the translation probabilities from the NMT model.

In iteration 2, the above process is repeated, but the synthetic training data are re-generated with the updated NMT models $M_{x\rightarrow y}^1$ and $M_{y\rightarrow x}^1$, which are presumably more accurate. 
In turn, the learnt NMT models $M_{x\rightarrow y}^2$ and $M_{y\rightarrow x}^2$ are also expected to be improved over the first iteration. 

The formal algorithm is listed in Algorithm~\ref{alg:jointTraining}, which is divided into two major steps: pre-training and joint training. As we will show in next section, the joint training step essentially adds an EM (Expectation-Maximization) process over the monolingual data in both source and target languages\footnote{Note that the training criteria on parallel data $D$ are still using MLE (maximum likelihood estimation) }.

\section{Training Objective}

Next we will show how to derive our new learning objective for joint training, starting with the case that only one NMT model is involved.

Given parallel corpus $D=\{(x^{(n)},y^{(n)})\}_{n=1}^{N}$ and monolingual corpus in target language $Y = \{y^{(t)}\}_{t=1}^{T}$, the semi-supervised training objective is to maximize the likelihood of both bilingual data and monolingual data:
\begin{equation}
L^*(\theta_{x\rightarrow y}) = \sum_{n=1}^N \log p(y^{(n)}|x^{(n)}) + \sum_{t=1}^T\log p(y^{(t)})
\label{equ:semi-loss}
\end{equation}
where the first term on the right side denotes the likelihood of bilingual data and the second term represents the likelihood of target-side monolingual data.
Next we introduce the source translations as hidden states for the target sentences and decompose $\log p(y^{(t)})$ as 
\begin{equation}
\begin{aligned}
\log p(y^{(t)}) & = \log \sum_x p(x,y^{(t)}) = \log \sum_x Q(x) \frac{p(x,y^{(t)})}{Q(x)}  \\
    & \geq \sum_x Q(x) \log \frac{p(x,y^{(t)})}{Q(x)} (\text{Jensen's inequality}) \\
    & = \sum_x [Q(x)\log p(y^{(t)}|x) - KL(Q(x)||p(x))]
\end{aligned}
\label{equ:loss}
\end{equation}
where $x$ is latent variable representing the source translation of target sentence $y^{(t)}$, $Q(x)$ is the approximated probability distribution of $x$, $p(x)$ represents the marginal distribution of sentence $x$, and $KL(Q(x)||p(x))$ is the Kullback-Leibler Divergence between two probability distributions. 
In order to make the equal sign to be valid in Equation \ref{equ:loss}, $Q(x)$ must satisfy the following condition
\begin{equation}
\frac{p(x,y^{(t)})}{Q(x)} = c
\end{equation}
where $c$ is a constant and does not depend on $y$. 
Given $\sum_x Q(x) = 1$, $Q(x)$ can be calculated as 
\begin{equation}
\begin{aligned}
Q(x) & = \frac{ p(x,y^{(t)}) }{c} = \frac{p(x,y^{(t)}) }{\sum_x p(x,y^{(t)}) } = p^*(x|y^{(t)}) 
\end{aligned}
\end{equation}
where $p^*(x|y^{(t)})$ denotes the true target-to-source translation probability.
Since it is usually not possible to calculate $p^*(x|y^{(t)})$ in practice, we use the translation probability $p(x|y^{(t)})$ given by a target-to-source NMT model as $Q(x)$.
Combining Equation~\ref{equ:semi-loss} and~\ref{equ:loss}, we have 
\begin{equation}
\begin{aligned}
& L^*(\theta_{x\rightarrow y}) \geq L(\theta_{x\rightarrow y}) = \sum_{n=1}^N \log p(y^{(n)}|x^{(n)}) + \\
&	 \sum_{t=1}^T\sum_x [p(x|y^{(t)}) \log p(y^{(t)}|x) - KL(p(x|y^{(t)})||p(x))]
\end{aligned}
\end{equation}
This means $L(\theta_{x\rightarrow y})$ is a lower bound of the true likelihood function $L^*(\theta_{x\rightarrow y})$.
Since $KL(p(x|y^{(t)})||p(x))$ is irrelevant to parameters $\theta_{x\rightarrow y}$, $L(\theta_{x\rightarrow y})$ can be simplified as
\begin{equation}
\begin{aligned}
L(\theta_{x\rightarrow y}) = & \sum_{n=1}^N \log p(y^{(n)}|x^{(n)}) \\
					& + \sum_{t=1}^T\sum_x p(x|y^{(t)}) \log p(y^{(t)}|x)
\end{aligned}
\label{equ:loss_x2y}
\end{equation}

The first part of $L(\theta_{x\rightarrow y})$ is the same as the MLE training, while the second part can be optimized with EM algorithm. We can estimate the expectation of source translation probability $p(x|y^{(t)})$ in the E-step, and maximize the second part in the M-step. 
The E-step uses the target-to-source translation model $M_{y\rightarrow x}$ to generate the source translations as hidden variables, which are paired with the target sentences to build a new distribution of training data together with true parallel data $D$. 
Therefore maximizing $L(\theta_{x\rightarrow y})$ can be approximated by maximizing the log likelihood on the new training data. 
The translation probability $p(x|y^{(t)})$ is used as the weight of the pseudo sentence pairs, which helps with filtering out bad translations.

It is easy to verify that back-translation approach \cite{Sennrich2016ImprovingNM} is a special case of this formulation of $L(\theta_{x\rightarrow y})$, in which $p(x|y^{(t)})$ = 1 because only the best translation from the NMT model $M_{y\rightarrow x}(y^{(t)})$ is used 
\begin{equation}
\begin{aligned}
L(\theta_{x\rightarrow y}) = & \sum_{n=1}^N \log p(y^{(n)}|x^{(n)}) 
\\ & + \sum_{t=1}^T \log p(y^{(t)}|M_{y\rightarrow x}(y^{(t)}))
\end{aligned}
\label{equ:loss_x2y_spe}
\end{equation}

Similarly, the likelihood of NMT model $M_{y\rightarrow x}$ can be derived as
\begin{equation}
\begin{aligned}
L(\theta_{y\rightarrow x}) = & \sum_{n=1}^N \log p(x^{(n)}|y^{(n)}) \\
					& + \sum_{s=1}^S\sum_y p(y|x^{(s)}) \log p(x^{(s)}|y)
\end{aligned}
\label{equ:loss_y2x}
\end{equation}
where $y$ is a target translation (hidden state) of the source sentence $x^{(s)}$. 
The overall training objective is the sum of likelihood in both directions
$$ L(\theta) = L(\theta_{x\rightarrow y}) + L(\theta_{y\rightarrow x}) $$

During the derivation of $L(\theta_{x\rightarrow y})$, we use the translation probability $p(x|y^{(t)})$ from $M_{y\rightarrow x}$ as the approximation of the true distribution $p^*(x|y^{(t)})$. 
When $p(x|y^{(t)})$ gets closer to $p^*(x|y^{(t)})$, we can get a tighter lower bound of $L^*(\theta_{x\rightarrow y})$, gaining more opportunities to improve $M_{x\rightarrow y}$. 
Joint training of paired NMT models is designed to solve this problem if source monolingual data are also available. 

\section{Experiments}

\subsection{Setup}
We evaluate our proposed approach on two language pairs: Chinese$\leftrightarrow$English and English$\leftrightarrow$German. 
In all experiments, we use BLEU~\cite{papineni2002bleu} as the evaluation metric for translation quality.

\begin{table*}[t]
	\centering
    \begin{tabular}{c|c|c||c|c|c|c|c}
    \hline
       Direction              &  System   & NIST2006 & NIST2003 & NIST2005 & NIST2008 & NIST2012 & Average \\ \hline \hline
    \multirow{4}{*}{C$\rightarrow$E}   & RNNSearch    & 38.61    & 39.39    & 38.31    & 30.04    & 28.48    & 34.97   \\
    								   & RNNSearch+M  & 40.66    & 43.26    & 41.61    & 32.48    & 31.16    & 37.83   \\ 
    								   & SS-NMT       & 41.53    & 44.03    & 42.24    & 33.40    & 31.58    &  38.56 \\ 
                                       & JT-NMT       & \textbf{42.56} & \textbf{45.10} & \textbf{44.36} & \textbf{34.10} & \textbf{32.26} & \textbf{39.67} \\ \hline
    \multirow{4}{*}{E$\rightarrow$C}   & RNNSearch    & 17.75    & 18.37    & 17.10    & 13.14    & 12.85    & 15.84   \\
                                       & RNNSearch+M  & 21.28    & 21.19    & 19.53    & 16.47    & 15.86    & 18.87   \\ 
                                       & SS-NMT       & 21.62    & 22.00    & 19.70    & 17.06    & 16.48    &  19.37  \\  
                                       & JT-NMT       & \textbf{22.56} & \textbf{22.98} & \textbf{20.95} & \textbf{17.62} & \textbf{17.39} & \textbf{20.30} \\ \hline
    \end{tabular}
	\caption{Case-insensitive BLEU scores (\%) on Chinese$\leftrightarrow$English translation. The ``Average" denotes the average BLEU score of all datasets in the same setting. The ``C" and ``E" denote Chinese and English respectively.}
	\label{table:chinese-english}
\end{table*}

\subsubsection{Dataset}
For Chinese$\leftrightarrow$English translation, we select our training data from LDC corpora\footnote{
The corpora include LDC2002E17, LDC2002E18, LDC2003E07, LDC2003E14, LDC2005E83, LDC2005T06, LDC2005T10,
LDC2006E17, LDC2006E26, LDC2006E34, LDC2006E85, LDC2006E92, LDC2006T06, LDC2004T08, LDC2005T10}, which consists of 2.6M sentence pairs with 65.1M Chinese words and 67.1M English words respectively.
We use 8M Chinese sentences and 8M English sentences randomly extracted from Xinhua portion of Gigaword corpus as the monolingual data sets.
Any sentence longer than 60 words is removed from training data (both the bilingual data and pseudo bilingual data).
For Chinese-English, NIST OpenMT 2006 evaluation set is used as validation set, and NIST 2003, NIST 2005, NIST 2008, NIST2012 datasets as test sets.
In both validation and test data sets, each Chinese sentence has four reference translations.
For English-Chinese, we use the NIST datasets in a reverse direction: treating the first English sentence in the four reference translation as a source sentence and the Chinese sentence as the single reference.
We limit the vocabulary to contain up to 50K most frequent words on both the source and target side, and convert remaining words into the {\tt <unk>} token.

For English$\leftrightarrow$German translation, we choose the WMT'14 training corpus used in ~\citet{jean-EtAl:2015:ACL-IJCNLP}.
This training corpus contains 4.5M sentence pairs with 116M English words and 110M German words.
For monolingual data, we randomly select 8M English sentences and 8M German sentences from ``News Crawl: articles from 2012" provided by WMT'14. 
The concatenation of news-test 2012 and news-test 2013 is used as the validation set and news-test 2014 as the test set.
The maximal sentence length is also set as 60.
We use 50K sub-word tokens as vocabulary based on Byte Pair Encoding~\cite{Sennrich2016NeuralMT}. 

\subsubsection{Implementation Details}
The RNNSearch model proposed by~\citet{Bahdanau2014NeuralMT} is adopted as our baseline, 
which uses a single layer GRU-RNN for the encoder and another.
The size of word embedding (for both source and target words) is 256 and the size of hidden layer is set to 1024.
The parameters are initialized using a normal distribution with a mean of 0 and a variance of $\sqrt{6/(d_{row}+d_{col})}$, where $d_{row}$ and $d_{col}$ are the number of rows and columns in the structure~\cite{glorot2010understanding}.
Our models are optimized with the Adadelta~\cite{zeiler2012adadelta} algorithm with mini-batch size 128.
We re-normalize gradient if its norm is larger than 2.0~\cite{pascanu2013difficulty}.
At test time, beam search with size 8 is employed to find the best translation,
and translation probabilities are normalized by the length of the translation sentences.
In post-processing step, we follow the work of~\citet{LuongACL2015} to handle {\tt <unk>} replacement for Chinese$\leftrightarrow$English translation.

For building the synthetic bilingual data in our approach, beam size is set to 4 to speed up the decoding process.
In practice, we first sort all monolingual data according to the sentence length and then 64 sentences are simultaneously translated with parallel decoding implementation.
As for model training, we found that 4-5 EM iterations are enough to converge.
The best model is selected according to the BLEU scores on the validation set during EM process.

\subsubsection{Baseline}
Our proposed joint-training approach is compared with three NMT baselines for all translation tasks:
\begin{itemize}
	\item 
	{\bf RNNSearch}: Attention-based NMT system~\cite{Bahdanau2014NeuralMT}.
    Only bilingual corpora are used to train a standard attention-based NMT model. 
    \item
	{\bf RNNSearch+M}: Bilingual and target-side monolingual corpora are used to train RNNSearch. 
   	We follow~\citet{Sennrich2016NeuralMT} to construct pseudo-parallel corpora by generating source language with back-translation of target-side monolingual data. 
    \item
    {\bf SS-NMT}: Semi-supervised NMT training proposed by \citet{Cheng2016SemiSupervisedLF}. To be fair in all experiment, their method adopts the same settings as our approach including the same source and target monolingual data.
\end{itemize}

\begin{table*}[t]
\centering
\begin{tabular}{l|l|c|c}
\hline
System                               & Architecture                                          & E$\rightarrow$D      & D$\rightarrow$E      \\ \hline \hline
\citet{jean-EtAl:2015:ACL-IJCNLP}    &   Gated RNN with search + PosUnk                      & 18.97                & -                    \\ \hline
\citet{jean-EtAl:2015:ACL-IJCNLP}    &   Gated RNN with search + PosUnk + 500K vocabs        & 19.40                & -                    \\ \hline
\citet{shen-EtAl:2016:P16-1}        & Gated RNN with search + PosUnk + MRT                   & 20.45                & -                    \\ \hline
\citet{luong2015effective}          &   LSTM with 4 layers + dropout + local att. + PosUnk   & 20.90                & -                    \\ \hline \hline
RNNSearch                           &   Gated RNN with search + BPE                          & 19.78                & 24.91                \\ \hline
RNNSearch+M                         &   Gated RNN with search + BPE + monolingual data       & 21.89                & 26.81                \\ \hline
SS-NMT                              &   Gated RNN with search + BPE + monolingual data       & 22.64                & 27.30                \\ \hline
JT-NMT                              &   Gated RNN with search + BPE + monolingual data       & \textbf{23.60}       & \textbf{27.98}       \\ \hline
\end{tabular}
\caption{Case-sensitive BLEU scores (\%) on English$\leftrightarrow$German translation. ``PosUnk" denotes~\citet{LuongACL2015}'s technique of handling rare words. ``MRT" denotes minimum risk training proposed in~\citet{shen-EtAl:2016:P16-1}. ``BPE" denotes Byte Pair Encoding proposed by~\citet{Sennrich2016NeuralMT} for word segmentation. The ``D" and ``E" denote German and English respectively. }
\label{table:english-german}
\end{table*}

\subsection{Chinese$\leftrightarrow$English Translation Result}

Table \ref{table:chinese-english} shows the evaluation results of different models on NIST datasets, in which JT-NMT represents our joint training for NMT using monolingual data.
All the results are reported based on case-insensitive BLEU.

Compared with RNNSearch, we can see that RNNSearch+M, SS-NMT and JT-NMT all bring significant improvements across different test sets.
Our approach achieves the best result, 4.7 and 4.46 BLEU points improvement over RNNSearch on average for Chinese-to-English and English-to-Chinese respectively.  
These results confirm that exploiting massive monolingual corpora improves translation performance.
 
From Table \ref{table:chinese-english}, we can find our JT-NMT achieves better performances than RNNSearch+M across different test sets, with 1.84 and 1.43 points of BLEU improvements on average in Chinese-to-English and English-to-Chinese directions respectively. Compared with RNNSearch+M, our joint training approach introduces data weight to better handle poor pseudo-training data, and the joint interactive training can boost the models of two directions with the help of each other, instead of only use the target-to-source model to help source-to-target model.  
Our approach also yields better translation than SS-NMT with at least 0.93 points BLEU improvements on average. 
This result shows that our method can better make use of both source and target monolingual corpora than \citet{Cheng2016SemiSupervisedLF}'s approach. 

\subsection{English$\leftrightarrow$German Translation Result}

For English$\leftrightarrow$German translation task, in addition to the baseline system, we also include results of other existing NMT systems, including  \citet{jean-EtAl:2015:ACL-IJCNLP}, \citet{shen-EtAl:2016:P16-1} and \citet{luong2015effective}.
In order to be comparable with other work, all the results are reported based on case-sensitive BLEU. Experiment results are shown in Table \ref{table:english-german}.

We can observe that the baseline RNNSearch with BPE method achieves better results than~\citet{jean-EtAl:2015:ACL-IJCNLP}, even better than the result using larger vocabulary of size 500K.
Compared with RNNSearch, we observe that RNNSearch+M, SS-NMT and JT-NMT bring significant improvements in both English-to-German and German-to-English directions.
It confirms the effectiveness of leveraging monolingual corpus.
Our approach outperforms RNNSearch+M and SS-NMT by a notable margin and obtains the best BLEU score of 23.6 and 27.98 in English-to-German and German-to-English test set respectively.
These experimental results further confirm the effectiveness of our joint training mechanism, similar as shown in the Chinese$\leftrightarrow$English translation tasks.

\subsection{Effect of Joint Training}

\begin{figure*}[t]
\centering
\begin{subfigure}[b]{0.24\textwidth}
\centering
\includegraphics[scale=0.23]{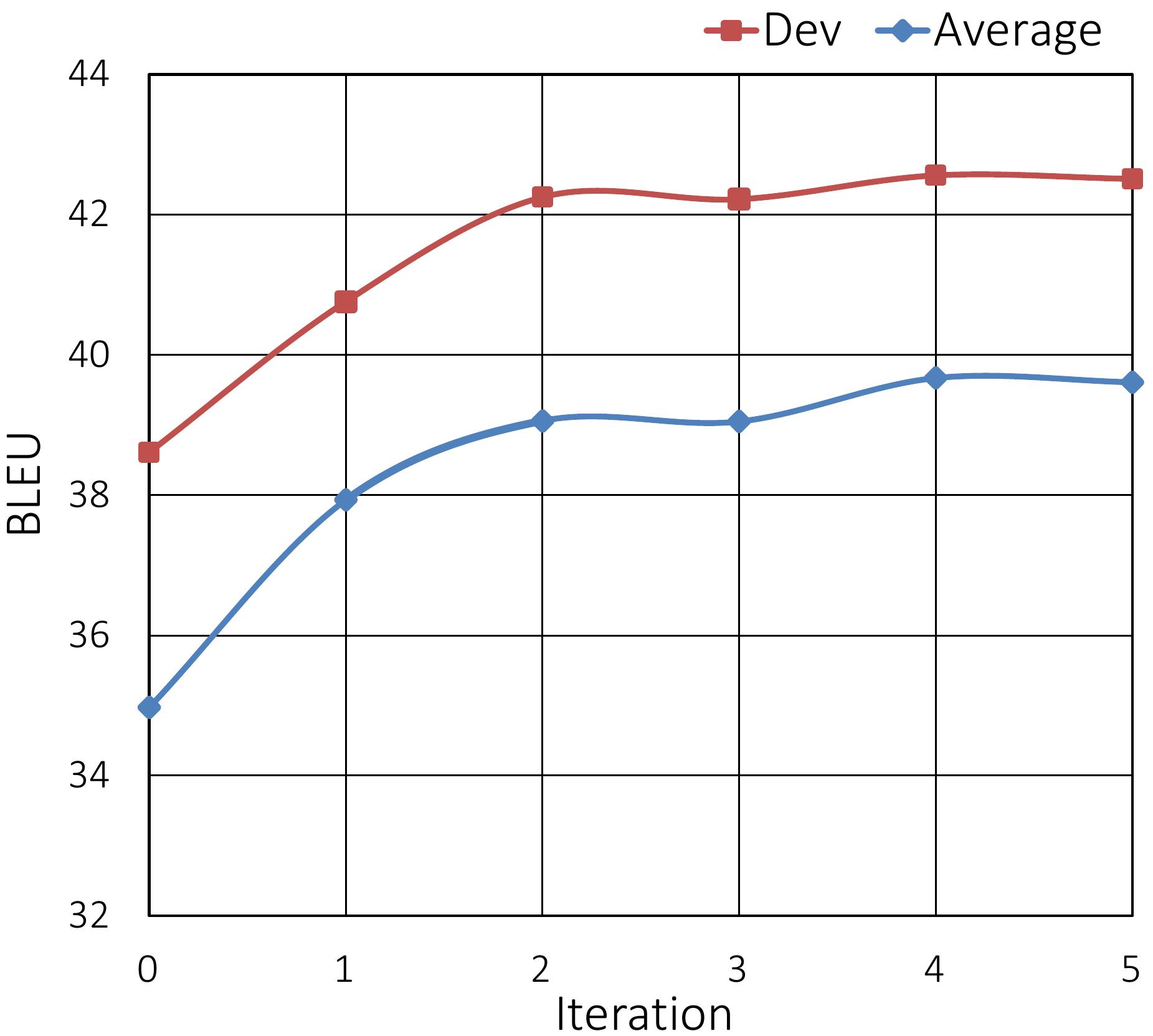}
\caption[Network2]%
{{\small Chinese-English Translation}}    
\label{fig:zh2en-curve}
\end{subfigure}
\hfill
\begin{subfigure}[b]{0.24\textwidth}  
\centering 
\includegraphics[scale=0.23]{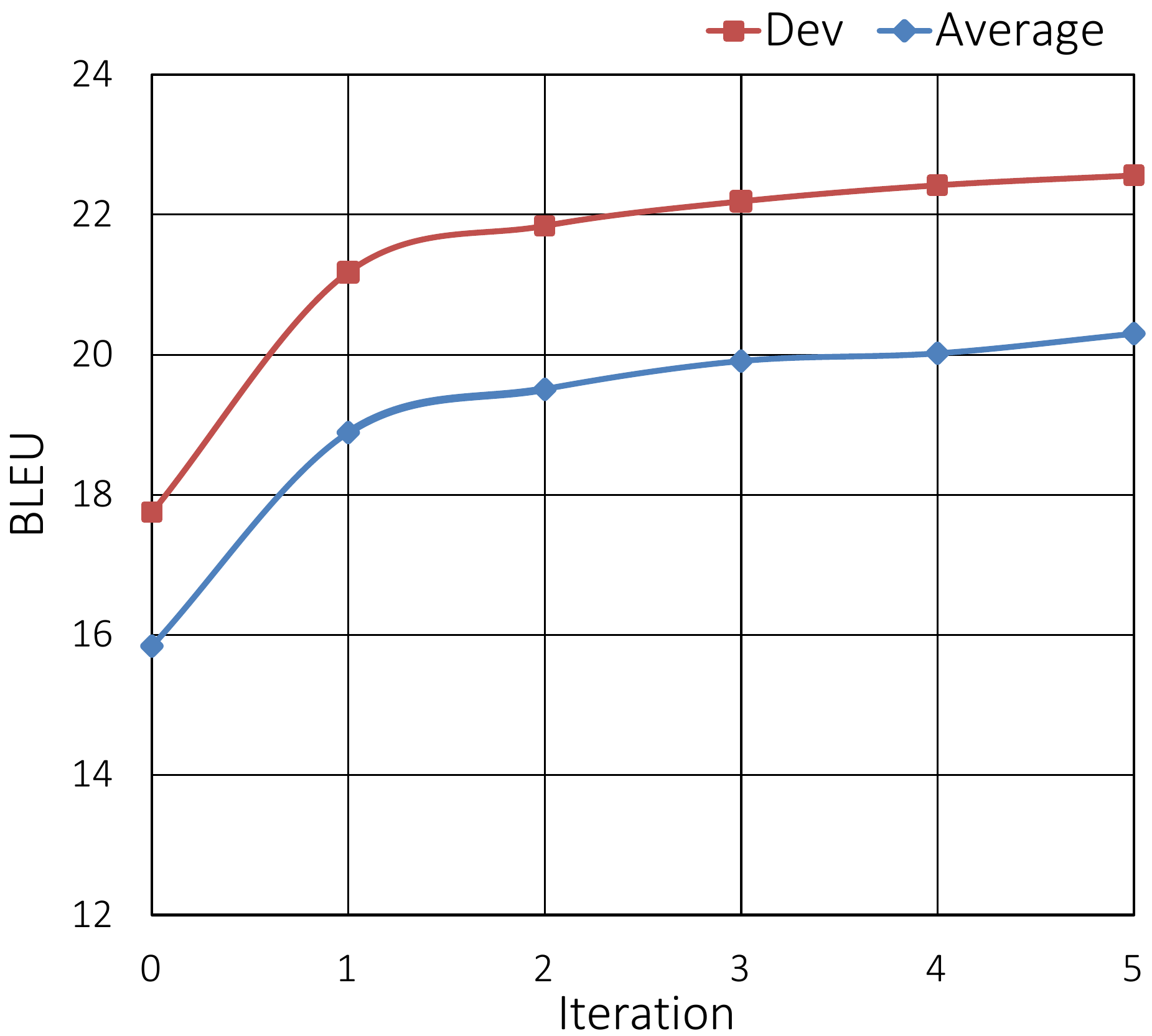}
\caption[]%
{{\small English-Chinese Translation}}    
\label{fig:en2zh-curve}
\end{subfigure}
\hfill
\begin{subfigure}[b]{0.24\textwidth}   
\centering 
\includegraphics[scale=0.23]{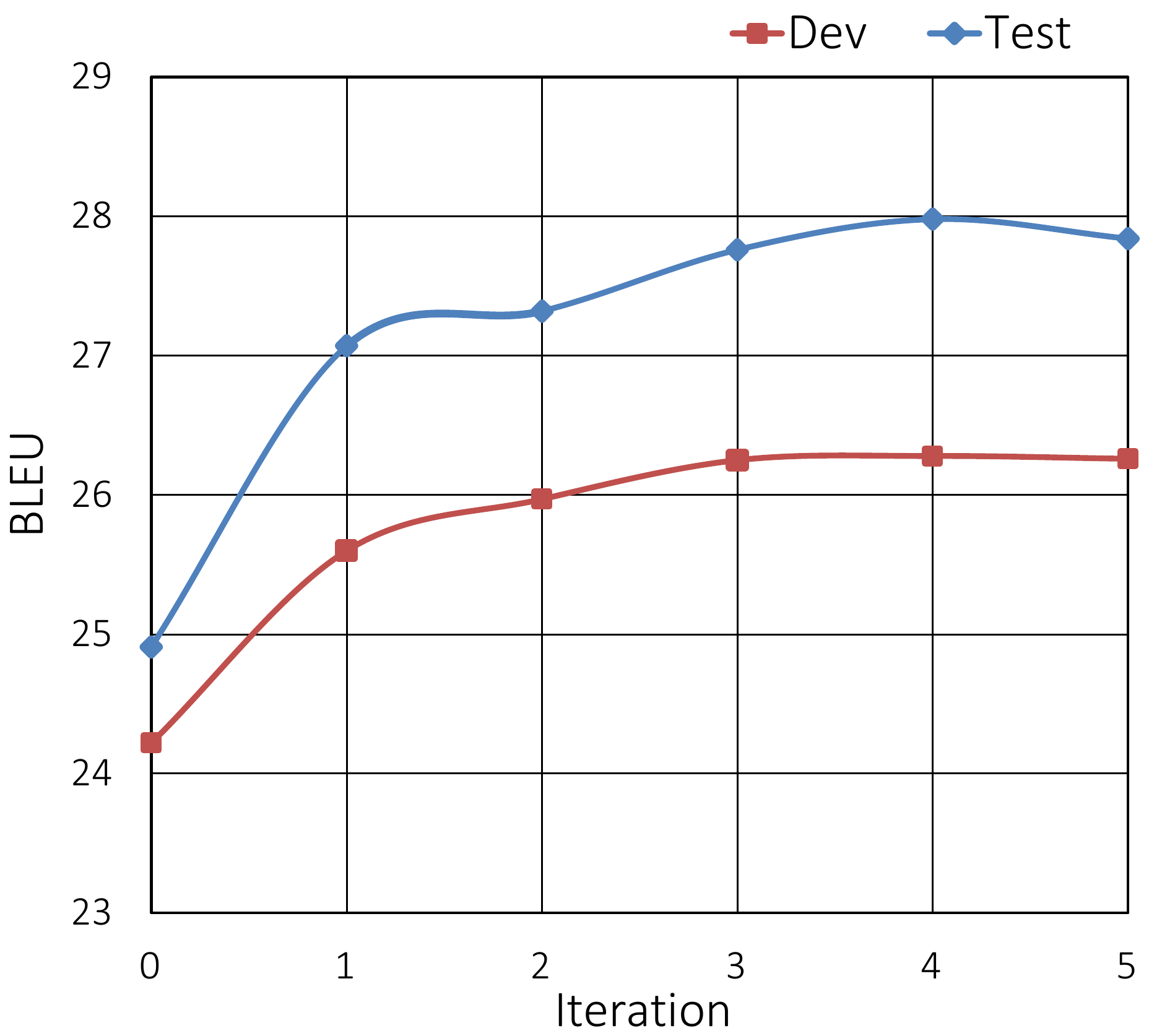}
\caption[]%
{{\small German-English Translation}}    
\label{fig:de2en-curve}
\end{subfigure}
\hfill
\begin{subfigure}[b]{0.24\textwidth}   
\centering 
\includegraphics[scale=0.23]{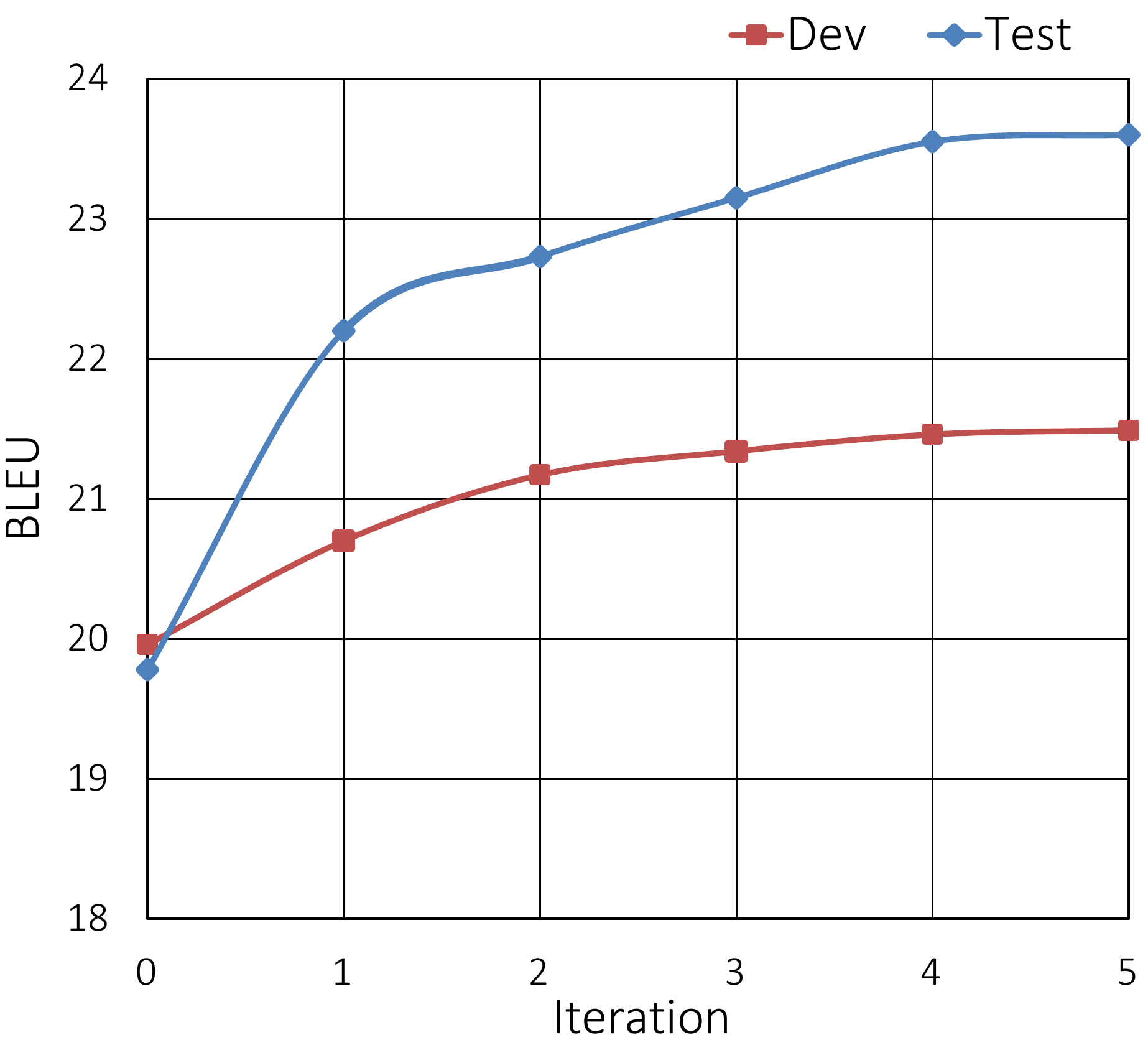}
\caption[]%
{{\small English-German Translation}}    
\label{fig:en2de-curve}
\end{subfigure}
\caption[]
{BLEU scores (\%) on Chinese$\leftrightarrow$English and  English$\leftrightarrow$German validation and test sets for JT-NMT during training process. ``Dev'' denotes the results of validation datasets, while ``Test'' denotes the results of test datasets. } 
\label{fig:effectOfJointTrain}
\end{figure*}

We further investigate the impact of our joint training approach JT-NMT during the whole training process.
Figure \ref{fig:effectOfJointTrain} shows the BLEU scores on Chinese$\leftrightarrow$English and English$\leftrightarrow$German validation and test sets in each iteration.
We can find that more iterations can lead to better evaluation results consistently, which verifies that the joint training of NMT models in two directions can boost their translation performance. 

In Figure \ref{fig:effectOfJointTrain}, ``Iteration 0'' is the BLEU scores of baseline RNNSearch, and obviously the first few iterations gain most, especially for ``Iteration 1''. After three iterations, we cannot get significant improvement anymore.
As we said previously, along with the target-to-source model approaches the ideal translation probability, the lower bound of the loss will be closer to the true loss. During the training, the closer the lower bound to the true loss, the smaller the potential gain. Since there is a lot of uncertainty during the training, the performance sometimes drops a little.

JT-NMT (Iteration 1) can be considered as the general version of RNNSearch+M that any pseudo sentence pair is weighted as 1. From Table \ref{table:sampleweight}, we can see that JT-NMT (Iteration 1) slightly surpass RNNSearch+M on all test datasets, which proves that the weight introduced in our algorithm can clean poor synthetic data and lead to better performance.
Our approach will assign low weight to synthetic sentence pairs with poor translation, so as to punish its effect to the model update. The translation will be refined and improved in subsequent iterations, as shown in Table~\ref{table:example}, which shows translation results of a Chinese sentence in different iterations. 

\begin{table}[t]
\centering
\caption{The BLEU scores (\%) on Chinese$\leftrightarrow$English and  English$\leftrightarrow$German translation tasks. For Chinese$\leftrightarrow$English translation, we list the average results of all test sets. For English$\leftrightarrow$German translation, we list the results of news-test2014.}
\begin{tabular}{c|c|c|c|c}
\hline
       System     & C$\rightarrow$E & E$\rightarrow$C    & D$\rightarrow$E    & E$\rightarrow$D    \\  \hline \hline
RNNSearch+M & 37.83           & 18.87 & 26.81 & 21.89 \\ \hline
JT-NMT (Iteration 1) & \textbf{38.23}           & \textbf{19.10} & \textbf{27.07} &\textbf{22.20} \\ \hline
\end{tabular}
\label{table:sampleweight}
\end{table}


\begin{table*}[t]
\centering
\begin{tabular}{l|l}
\hline
Monolingual                  & \makecell[l]{当~终场~哨声~响~起~,~意大利~首都~罗马~沸腾~了~。 \\
\textit{dang zhongchang shaosheng xiang qi , yidali shoudu luoma feiteng le .}}\\
\hline
Reference                    & \makecell[l]{ {\color{blue} when the final whistle sounded }, the italian capital of rome boiled . }\\ 
\hline
\multirow{3}{*}{Translation} & \makecell[l]{{[}Iteration 0{]}: the italian capital of rome was boiling {\color{blue} with the rome }.  } \\ \cline{2-2} 
                             & \makecell[l]{{[}Iteration 1{]}: the italian capital of rome was boiling {\color{blue} with the sound of the end} {\color{blue} of the door } .} \\ \cline{2-2} 
                             & \makecell[l]{{[}Iteration 4{]}: {\color{blue} when the final whistle} {\color{blue} sounded }, the italian capital of rome was boiling .} \\ 
\hline
\end{tabular}
\caption{Example translations of a Chinese sentence in different iterations.}
\label{table:example}
\end{table*}

\section{Related Work}
Neural machine translation has drawn more and more attention in recent years \cite{Bahdanau2014NeuralMT,luong2015effective,jean-EtAl:2015:ACL-IJCNLP,tu-EtAl:2016:P16-1,Wu2016GooglesNM}.
For the original NMT system, only parallel corpora can be used for model training using MLE method, therefore much research in the literature attempts to exploit massive monolingual corpora.
\citet{gulcehre2015using} first investigate the integration of monolingual data for neural machine translation.
They train monolingual language models independently, which is integrated into the NMT system with proposed shallow and deep fusion methods.
\citet{Sennrich2016ImprovingNM} propose to generate the synthetic bilingual data by translating the target monolingual sentences to source language sentences, and the mixture of original bilingual data and the synthetic parallel data are used to retrain the NMT system.
As an extension of their approach, our approach introduces translation probabilities from target-to-source model as weights of synthetic parallel sentences to punish poor pseudo parallel sentences, and further interactive training of NMT models in two directions are used to refine them.

Recently, \citet{zhang2016exploiting} propose a multi-task learning framework to exploit source-side monolingual data, in which they jointly perform machine translation on synthetic bilingual data and sentence reordering with source-side monolingual data.
\citet{Cheng2016SemiSupervisedLF} reconstruct monolingual data by auto-encoder, in which the source-to-target and target-to-source translation models form a closed loop and are jointly updated.
Different from their method, our approach extends \citet{Sennrich2016ImprovingNM} by directly introducing source-side monolingual data to improve reverse NMT models and adopts EM algorithm to iteratively update bidirectional NMT models.
Our approach can better exploit both target and source monolingual data, while they show no improvement when using both target and source monolingual data compared just target monolingual data.
\citet{he2016dual} treat the source-to-target and target-to-source models as the primal and dual tasks respectively, similar to the work of \citet{Cheng2016SemiSupervisedLF}, they also employed round-trip translations for each monolingual sentence to obtain feedback signals. 
\citet{ramachandran-liu-le} adopt pre-trained weights of two language models to initial the encoder and decoder of a seq2seq model, and then fine-tune it with labeled data. 
Their approach is complementary to our mechanism by leveraging pre-trained language model to initial bidirectional NMT models, and it may lead to additional gains.

\section{Conclusion}

In this paper, we propose a new semi-supervised training approach to integrating the training of a pair of translation models in a unified learning process with the help of monolingual data from both source and target sides.
In our method, a joint-EM training algorithm is employed to optimize two translation models cooperatively, in which the two models are able to mutually boost their translation performance. 
Translation probability of the other model is used as the weight to estimate translation accuracy and punish the bad translations.
Empirical evaluations are conducted in Chinese$\leftrightarrow$English and English$\leftrightarrow$German translation tasks, and demonstrate that our approach leads to significant improvements, compared with strong baseline systems. In the future work, we plan to extend this method to jointly train multiple NMT systems for 3+ languages using massive monolingual data. 

\section*{Acknowledgments}
This research was partially supported by grants from the National Natural Science Foundation of China (Grants No. 61727809, 61325010 and U1605251).
We appreciate Dongdong Zhang, Shuangzhi Wu, Wenhu Chen, Guanlin Li for the fruitful discussions. 
We also thank the anonymous reviewers for their careful reading of our paper and insightful comments.

\end{CJK*}

\bibliographystyle{aaai}
\bibliography{aaai}

\end{document}